\documentclass{article}
\usepackage[utf8]{inputenc} 
\usepackage[T1]{fontenc}    
\usepackage{hyperref}       
\usepackage{url}            
\usepackage{booktabs}       
\usepackage{amsfonts}       
\usepackage{nicefrac}       
\usepackage{microtype}      
\usepackage{lipsum}         
\usepackage{fancyhdr}       
\usepackage{graphicx}       
\usepackage{xcolor}
\graphicspath{{media/}}  
\usepackage{amsmath}

\title{Multi-Task Learning for Affect Analysis}

\author{
  Fazeel Asim \\
}

\begin{document}
\maketitle

\begin{abstract}
This research delves into the realm of affective computing for image analysis, aiming to
enhance the efficiency and effectiveness of multi-task learning in the context of emotion
recognition.

This project investigates two primary approaches: uni-task solutions and a multi-task
approach to the same problems. Each approach undergoes testing, exploring various
formulations, variations, and initialization strategies to come up with the best configuration.
The project utilizes existing a neural network architecture, adapting it for multi-task learning
by modifying output layers and loss functions. Tasks encompass 7 basic emotion
recognition, action unit detection, and valence-arousal estimation. Comparative analyses
involve uni-task models for each individual task, facilitating the assessment of multi-task
model performance.

Variations within each approach, including, loss functions, and hyperparameter tuning,
undergo evaluation. The impact of different initialization strategies and pre-training
techniques on model convergence and accuracy is explored. The research aspires to
contribute to the burgeoning field of affective computing, with applications spanning
healthcare, marketing, and human-computer interaction.
By systematically exploring multi-task learning formulations, this research aims to contribute
to the development of more accurate and efficient models for recognizing and
understanding emotions in images. The findings hold promise for applications in diverse
industries, paving the way for advancements in affective computing.
\end{abstract}

\section{Introduction}
In the age of rapid technological advancement, the realm of affective computing stands
at the forefront of innovation, promising insights into human emotions and interactions.
At the face of this domain lies the captivating field of image analysis for emotion
recognition, where the fusion of artificial intelligence and psychology seeks to decipher
the complexities of human affect.

The overarching goal of this research is twofold: to explore and evaluate the performance
of uni-task models and compare them against their multi-task counterparts. By
leveraging existing neural network architectures and modifying them for multi-task
learning, we aim to transcend the boundaries of conventional image analysis techniques.

\subsection{Background}
Affective computing, a multidisciplinary field at the intersection of computer science,
psychology, and artificial intelligence, has witnessed remarkable progress in recent
decades. The endeavour to introduce machines with the ability to comprehend and
respond to human emotions through image analysis. This holds substantial promise
across diverse domains. Applications range from healthcare diagnostics to personalized
user experiences, with potential implications in marketing and human-computer
interaction.

\subsection{Problem Statement}
While the potential of affective computing in image analysis is great, the accurate and
robust recognition of emotions in a multi-task setting remains a significant challenge.
Existing methodologies often face limitations in capturing the nuanced and complex
nature of emotional states. Identifying an optimal multi-task learning formulation,
alongside strategic initialization, and pre-training methods, is crucial for advancing the
capabilities of affective computing models.

\subsection{Aim}
This research aims to investigate and optimize multi-task learning formulations for
affective computing in image analysis. Specifically, the focus is on enhancing the
simultaneous performance of basic emotion recognition, action unit detection, and
valence-arousal estimation. By exploring variations in model architectures and
initialization strategies, we aim to contribute to the development of more accurate and
efficient models in affective computing.

\subsection{Objectives}
\begin{itemize}
    \item Explore and compare the standard and alternative approaches to multi-task learning in affective computing.
    \item Investigate variations within each approach, including network architectures and loss functions.
    \item Evaluate the impact of different initialization and pre-training strategies on model performance.
    \item Create and compare uni-task models for basic emotion recognition, action unit detection, and valence-arousal estimation.
    \item Identify the most effective multi-task learning formulation and initialization strategy for enhanced affective computing.
\end{itemize}

\subsection{Research Questions}
\begin{itemize}
    \item How do different variations within each approach affect the performance of models?
    \item What is the impact of various initialization and pre-training strategies on the convergence speed and accuracy of affective computing models?
    \item How does the performance of multi-task models compare to uni-task models for basic emotion recognition, action unit detection, and valence-arousal estimation?
\end{itemize}

\subsection{Literature Review}

\subsubsection{Affective Computing and Image Analysis}
In recent years, affective computing has emerged as a prominent field, aiming to understand and respond to human emotions. Within image analysis, the recognition and interpretation of emotions from visual content offer significant application potential. Research in this area encompasses methodologies like facial expression analysis, action unit detection, and valence-arousal estimation. Some notable contributions include: the development of FaceBehaviorNet, which employs distribution matching for heterogeneous multi-task learning in large-scale face analysis, addressing tasks like continuous affect estimation, facial action unit detection, and basic emotion recognition; The introduction of FACERNET, a dynamic multi-output Facial Expression Intensity Estimation network, facilitating accurate emotion recognition from varying-length videos with video-level annotations; and the expansion of the Aff-Wild database to Aff-Wild2, enabling comprehensive studies on affective computing by incorporating continuous emotions, basic expressions, and action unit annotations, ultimately leading to state-of-the-art performance in emotion recognition tasks across diverse datasets.

\subsubsection{Affective Computing and Image Analysis}
In recent years, affective computing has emerged as a prominent field, aiming to understand and respond to human emotions. Within image analysis, the recognition and interpretation of emotions from visual content offer significant application potential. Research in this area encompasses methodologies like facial expression analysis, action unit detection, and valence-arousal estimation. Some notable contributions include: the development of FaceBehaviorNet, which employs distribution matching for heterogeneous multi-task learning in large-scale face analysis, addressing tasks like continuous affect estimation, facial action unit detection, and basic emotion recognition; The introduction of FACERNET, a dynamic multi-output Facial Expression Intensity Estimation network, facilitating accurate emotion recognition from varying-length videos with video-level annotations; and the expansion of the Aff-Wild database to Aff-Wild2, enabling comprehensive studies on affective computing by incorporating continuous emotions, basic expressions, and action unit annotations, ultimately leading to state-of-the-art performance in emotion recognition tasks across diverse datasets.

\subsubsection{Affective Computing and Image Analysis}
In recent years, affective computing has emerged as a prominent field, aiming to understand and respond to human emotions. Within image analysis, the recognition and interpretation of emotions from visual content offer significant application potential. Research in this area encompasses methodologies like facial expression analysis, action unit detection, and valence-arousal estimation. Some notable contributions include: the development of FaceBehaviorNet, which employs distribution matching for heterogeneous multi-task learning in large-scale face analysis, addressing tasks like continuous affect estimation, facial action unit detection, and basic emotion recognition; The introduction of FACERNET, a dynamic multi-output Facial Expression Intensity Estimation network, facilitating accurate emotion recognition from varying-length videos with video-level annotations; and the expansion of the Aff-Wild database to Aff-Wild2, enabling comprehensive studies on affective computing by incorporating continuous emotions, basic expressions, and action unit annotations, ultimately leading to state-of-the-art performance in emotion recognition tasks across diverse datasets.

\subsubsection{Basic Emotion Recognition}
Basic emotion recognition involves identifying and categorizing primary emotions
particularly, happiness, sadness, anger, fear, disgust, and surprise. Ekman and Friesen's
seminal work laid the foundation for understanding basic emotions. They identified
universal facial expressions, such as happiness, sadness, anger, fear, disgust, and
surprise, providing a standardized framework for emotion recognition across cultures.
 (Ekman \& Friesen, 1971). Researchers have explored various methodologies, including
facial expression analysis, physiological signals, and speech prosody, to develop robust
emotion recognition systems. Facial expression recognition has gained significant
attention due to its non-intrusive nature and widespread use of cameras in modern
devices. Recent studies have focused on improving the accuracy of basic emotion
recognition algorithms by incorporating deep learning techniques. Convolutional Neural
Networks (CNNs) and Recurrent Neural Networks (RNNs) have shown promising results
in capturing complex patterns in facial expressions and other modalities. Challenges in
real-world scenarios, such as variations in lighting conditions, facial occlusions, and
cultural differences, continue to be addressed to enhance the applicability of these
systems.

\begin{figure}[h]
    \centering
    \includegraphics[width=0.5\textwidth]{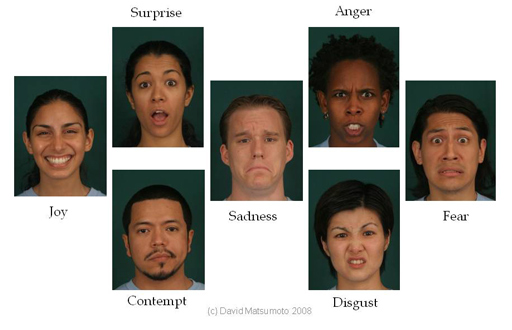}
    \caption{ Anger, Contempt, Fear, Disgust, Happiness, Sadness and Surprise.}
    \label{fig:example1}
\end{figure}

\subsubsection{Action Unit Detection}
Action unit detection involves the identification and analysis of specific facial muscle
movements associated with emotional expressions, providing a fine-grained
understanding of emotional states. Littlewort et al. (2004) introduced a method for
automatically extracting facial expression dynamics from video data, contributing to the
development of automated systems for action unit detection. Recent research combines
computer vision techniques with machine learning algorithms, integrating 3D facial
imaging, depth sensors, and landmark tracking. Deep learning architectures, such as
convolutional neural networks and recurrent neural networks, have shown promise in
automatically recognizing subtle facial movements indicative of specific emotional
expressions.

\begin{figure}[h]
    \centering
    \includegraphics[width=0.5\textwidth]{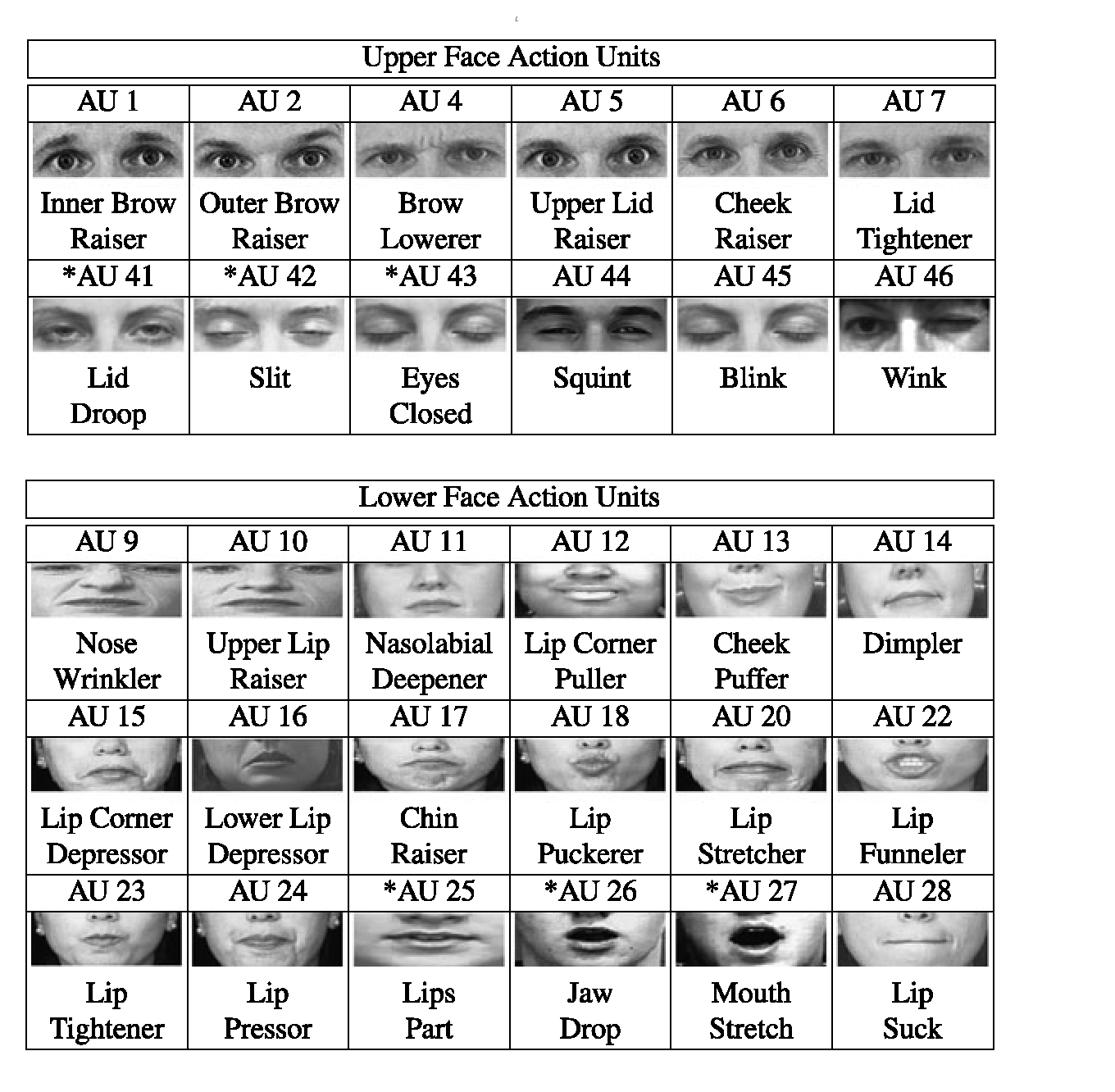}
    \caption{Some facial Action Units}
    \label{fig:example2}
\end{figure}

\subsubsection{Valence-Arousal Estimation}
Valence and arousal are dimensions that describe the emotional intensity and polarity of
an emotional experience. Valence ranges from positive to negative, while arousal
represents the level of emotional activation. Estimating valence and arousal is essential
for a more comprehensive understanding of emotional states. Russell's circumplex
model provides a theoretical framework for understanding emotions in terms of valence
and arousal (Russell, 1980). 

\begin{figure}[h]
    \centering
    \includegraphics[width=0.5\textwidth]{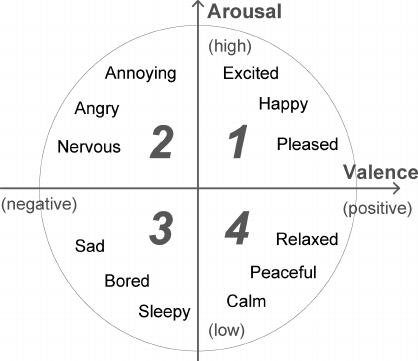}
    \caption{Valence and Arousal metrics in 2D space}
    \label{fig:example3}
\end{figure}

This model has influenced research on valence-arousal
estimation by providing a structured representation of emotional states. Machine learning
models, including regression and dimensional approaches, have been employed for
valence-arousal estimation. Integrating multiple modalities, such as facial expressions,
speech, and physiological signals, has been explored to enhance the accuracy of
predictions. Challenges include the subjective nature of emotional experiences,
individual differences in emotional expression, and the need for multimodal fusion
techniques.

\subsection{The Dataset}
In the pursuit of advancing affective computing methodologies, the selection of an
appropriate dataset is paramount. The dataset chosen for this project, Aff-Wild2 \cite{kollias2019deep,kollias2019expression,kollias2022abaw,kollias2023abaww,kollias2023btdnet,kollias2023facernet,kollias2023multi,kollias20246th,kollias2024distribution,kollias2024domain,kolliasijcv,zafeiriou2017aff,hu2024bridging,psaroudakis2022mixaugment,kollias2020analysing,kollias2021distribution,kollias2021affect,kollias2019face,kollias2021analysing,kollias2023ai,kollias2023deep2,arsenos2023data,gerogiannis2024covid},
emerges as a fundamental resource in our endeavour to explore the nuances of emotion
recognition within image analysis. Aff-Wild2, an extension of the renowned Aff-Wild
database, addresses critical limitations encountered in previous datasets, making it an
ideal choice for our research objectives.

Affective computing research has historically faced challenges stemming from the
scarcity of comprehensive in-the-wild datasets. Aff-Wild2 steps in to fill this void by
substantially expanding upon its predecessor, Aff-Wild, to encompass a broader scope
of annotations and task coverage. Unlike its predecessors and other existing datasets,
Aff-Wild2 offers a holistic solution, providing annotations for valence-arousal estimation,
action unit detection, and basic expression classification across a diverse array of
subjects and environmental contexts.

The decision to utilize Aff-Wild2 is driven by its unique attributes, which align closely with
the objectives of this project. Firstly, Aff-Wild2 features annotations for continuous
emotions, such as valence and arousal, enabling us to delve deeper into the dynamics
of human affect. Secondly, the dataset includes annotations for basic expressions and
action units, facilitating a comprehensive analysis of emotion recognition tasks. Thirdly,
Aff-Wild2 boasts a significant scale, comprising 564 videos totalling approximately 2.8
million frames, showcasing a rich diversity of subjects in terms of age, ethnicity, and
nationality, as well as environmental contexts.

In terms of annotation granularity, Aff-Wild2 stands out by offering per-frame annotations
for basic expressions, action units, and valence-arousal dimensions. This careful
annotation process ensures that the dataset captures the subtle nuances of facial
expressions and gestures, essential for training and evaluating robust emotion
recognition models.

Aff-Wild2 provides a solid foundation, offering real-world data to validate our
methodologies and algorithms. The dataset's comprehensive coverage, diverse subject
pool, and annotations position it as a foundation for this project of affective computing.
In summary, Aff-Wild2 emerges as an necessary asset in our project, offering a wealth
of data and annotations to fuel our exploration of emotion recognition in image analysis. 

\subsection{The Uni-Task Problems}
This chapter outlines the specifics of three distinct uni-task problems based on the AffWild2 database: the VA Estimation problem, the Expression Recognition problem, and
the AU Detection Challenge. Each challenge focuses on a specific aspect of affect
recognition, with unique opportunities to develop and evaluate specialized models.
However, for the purposes of this project, Resnet 18 will be used for all problems for
continuity and fair comparisons. All of the tasks are done in python programming in
Jupiter notebooks.

\subsubsection{Valence-Arousal Estimation}
This problem revolves around estimating valence and arousal from cropped and aligned
video data. The dataset provides with two sub-folders: Train Set and Validation Set,
containing annotation files corresponding to training and validation data. Each annotation
file follows a consistent format, with the first line indicating "valence, arousal" and
subsequent lines containing comma-separated valence and arousal values for each
video frame. It's important to note that valence and arousal values range from -1 to 1,
with -5 denoting disregarded frames.

For this task, performance is assessed based on the mean
Concordance Correlation Coefficient (CCC) of valence and arousal.

\begin{equation} \label{va}
\mathcal{P}_{VA} = \frac{\rho_a + \rho_v}{2},
\end{equation}

The figure above denotes the general evaluation criteria. 

The Concordance Correlation Coefficient (CCC) serves as the primary performance
measure, assessing the agreement between the annotations and predictions of valence
and arousal. CCC evaluates the correlation between two time series by scaling their
correlation coefficient with their mean square difference. It yields values in the range of
[-1, 1], with higher values indicating better performance.

\subsection{The Implementation}
The code implements a deep learning model for affective computing, specifically
targeting valence and arousal estimation in images. Affective computing aims to
understand and model human emotions using computational methods. The model
architecture leverages transfer learning with a pre-trained ResNet-18 convolutional
neural network (CNN), fine-tuning it for the specific task.

The dataset is crucial for training an effective model. This code utilizes a custom dataset
class (CustomDataset) to load images and their corresponding annotations. Annotations,
representing valence-arousal pairs, are extracted from annotation files provided in
specific folders. Images are pre-processed using data augmentation techniques such as
resizing, random horizontal flipping, and rotation. Data augmentation helps in increasing
the diversity of the training data, thereby improving the model's generalization ability.
Creating this Custom dataset alongside its data loaders was a tedious task as the
annotations were stored in text files and not in the regular csv format. This was my first
time I worked with annotations of this sort and what helped me a lot in creating it was the
Writing Custom Datasets, Data Loaders and Transforms tutorial by pytorch.

The custom dataset class (`CustomDataset`) plays a vital role in preparing the data for
training the affective computing model. It handles the loading of images and their
corresponding annotations, adhering to the requirements specified for the VA Estimation
dataset. Upon initialization, the class takes as input the paths to the annotation folder
and image folder. The `create custom dataset` method iterates through each
annotation file within the annotation folder, extracting valence-arousal pairs while
disregarding frames with invalid (-5) valence or arousal values, as per the challenge
requirements. For each valid annotation, it constructs the image path based on the video
name and frame number and checks if the corresponding image exists. If so, it adds the
video name, frame number, and annotations to the dataset. The `getitem` method
allows for retrieving a sample at a particular index, where it returns the image tensor and
annotations as a tensor. 

Additionally, the class employs data augmentation techniques,
including resizing, horizontal flipping, and rotation, within the transformation pipeline
specified during initialization. This ensures that the dataset is appropriately preprocessed and augmented for training the model, enhancing its ability to generalize to
unseen data. Through meticulous handling of annotations and images, the custom
dataset class ensures that the data fed into the model is well-prepared and aligned with
the requirements of the affective computing task.

The creation of this custom dataset comes in handy as the same variation of this custom
dataset is used for all the other problems too with specific changes applied on it to work
on different annotations.
To verify the validity of this custom dataset and the proper functioning of the data loaders,
I implemented a random image generator function that displays a random image and
alongside it prints information about the image. This information can be used to match
the image taken from the data loader to the corresponding annotation values. We can
cross reference these values with the actual ones on the system files

The Image correctly corresponds to the location, index, and annotation values for
valence and arousal. We can run this multiple times if needed and each time it will give
a different image alongside its annotations correctly. This function, in combination with
the custom dataset is used in all the other problems.
After this function I like to get the total number of images in both sets in which case it
comes out to be:
Total number of images in train set: 1653401
Total number of images in validation set: 374656
This is useful for calculating the number of samples we need for training.

Training and validation datasets are instantiated using the custom dataset class. The
total number of images in each set is printed for reference. Sample images with their
corresponding annotations are visualized to provide insight into the dataset's content and
structure. The code loads a pre-trained ResNet-18 model, a popular convolutional neural
network architecture, and modifies its fully connected layer to output two features
corresponding to valence and arousal. Batch normalization layers are inserted after
convolutional layers to improve model performance and stability. A training function is
defined to train the model using the provided datasets, loss function (Mean Squared
Error), optimizer (Adam optimizer), and learning rate scheduler (ReduceLROnPlateau
scheduler). This function iterates over epochs, performing forward and backward passes,
updating model parameters, and evaluating performance on the validation set.
The model is trained using the defined training function. Epoch-wise training and
validation losses are displayed, along with plots illustrating the training and validation
losses over epochs. Additionally, the Concordance Correlation Coefficient (CCC) for
valence and arousal is calculated and plotted to assess model performance.

The training process unfolds over 10 epochs, each characterized by adjustments in both
training and validation performance metrics. Initially, the model displays a relatively high
training loss of 0.0712, steadily decreasing across subsequent epochs. Simultaneously,
the validation loss experiences fluctuations, suggesting the model's ability to adapt to
unseen data but with occasional setbacks. Concordance Correlation Coefficients (CCC)
for valence and arousal are also tracked, showcasing the model's proficiency in capturing
emotional nuances within images. Notably, while the CCC values exhibit variability, they
generally maintain positive trends, indicating the model's capability to discern emotional
states to a certain degree. By the final epoch, the model achieves a more stabilized
performance, with a training loss of 0.0516 and a validation loss of 0.1239. This indicates
that the model has undergone iterative refinement, resulting in improved accuracy and
generalization. 

These are the results achieved from training on 0.8
Validation CCC Valence: 0.1652, Validation CCC Arousal: 0.3299
Average Validation CCC: 0.2476
The CCC could be improved a further by training it on more data but due to hardware
and time limitations, it wasn’t feasible to run it on the entire dataset.
Accuracy in the code is improved through several key mechanisms implemented
throughout the pipeline.

Firstly, data augmentation techniques are employed during training to increase the
variability and richness of the dataset. Transformations such as random horizontal
flipping and rotation are applied to the images, allowing the model to learn from a wider
range of visual representations. This augmentation helps prevent overfitting by
presenting the model with diverse perspectives of the same data, enhancing its ability to
generalize to unseen examples.

Secondly, batch normalization layers are inserted after convolutional layers in the model
architecture. Batch normalization normalizes the activations of each layer, effectively
stabilizing and accelerating the training process. By reducing internal covariate shift,
batch normalization allows for faster convergence and better utilization of the model's
capacity, ultimately leading to improved accuracy.
Furthermore, the choice of a pre-trained ResNet-18 model as the backbone architecture
provides a strong foundation for feature extraction. Pre-trained models are trained on
large-scale datasets, learning hierarchical features that are transferable to various tasks.
By fine-tuning this pre-trained model on the specific task of valence and arousal
estimation, the model can leverage the learned representations to achieve higher
accuracy with less training data.

Additionally, the use of the Mean Squared Error (MSE) loss function during training
contributes to accuracy improvement. MSE penalizes large errors more heavily than
small errors, encouraging the model to focus on minimizing the discrepancies between
predicted and ground truth values. This loss function is well-suited for regression tasks
like valence and arousal estimation, where the goal is to predict continuous values.
Lastly, the learning rate scheduler, specifically the ReduceLROnPlateau scheduler,
dynamically adjusts the learning rate during training based on the model's performance
on the validation set. This adaptive learning rate optimization technique ensures that
learning is appropriately adjusted as the training progresses, preventing overshooting or
stagnation in the optimization process. By fine-tuning the learning rate in response to
validation performance, the scheduler helps the model converge more effectively,
leading to improved accuracy.

\subsubsection{Expression Recognition Classification}
This problem concerns classifying facial expressions into predefined categories,
including neutral, anger, disgust, fear, happiness, sadness, surprise, and 'other'. Similar
to the VA Estimation Challenge, the dataset contains annotation files organized into
Train Set and Validation Set sub-folders. Each annotation file begins with a header line
listing the expression categories, followed by lines indicating the corresponding
expression annotation values for each video frame. Frames with annotation values of -1
should be disregarded.

For this part, a subset of the Aff-Wild2 database comprising 548 videos, totalling around
2.7 million frames, is utilized. These videos are annotated with annotations for the six
basic expressions (anger, disgust, fear, happiness, sadness, surprise), along with the
neutral state, and a category labelled 'other', denoting expressions or affective states
beyond the six basic ones.

The performance of models in this challenge is assessed based on the average F1 Score
across all eight categories, including the six basic expressions, the neutral state, and the
'other' category. The F1 score, a weighted average of precision and recall, serves as the
performance measure. A high F1 score, ranging from 0 to 1, is desirable, indicating
accurate classification performance.

The evaluation criterion for the Expression Recognition task is calculated as
follows: 

\begin{equation} \label{expr}
\mathcal{P}_{EXPR} = \frac{\sum_{expr} F_1^{expr}}{8}
\end{equation}

\subsection{The Implementation}

The script begins by importing necessary libraries and defining paths to image folders
and annotation files. It utilizes PyTorch's Dataset and DataLoader classes to create
custom datasets for training and validation. The CustomDataset class parses annotation
files to retrieve emotion labels and annotations for each frame, ensuring data consistency
and integrity.

Transformations such as resizing, random horizontal flipping, and random rotation are
applied to augment the training data, enhancing the model's generalization ability. After
preparing the datasets, the script provides functionality to explore and analyse the data.
It randomly samples images from either the training or validation set and displays
information about the sampled image, including video name, frame number, image
name, and annotations. Furthermore, it computes and visualizes the distribution of class
images in the dataset, facilitating an understanding of class imbalance and data
characteristics.

I create subsets of the training and validation datasets. This allows for quicker
experimentation and debugging without sacrificing the representativeness of the data.
The subsets are randomly sampled from the original datasets and used to train and
evaluate the model efficiently. The core of the script involves defining the deep learning
model architecture and training it on the provided datasets. A pre-trained ResNet-18
model is loaded and adapted by modifying its last layer to match the number of emotion
classes. The training loop iterates over multiple epochs, during which the model learns
to predict emotion labels from input images. To handle class imbalance, a weighted loss
function is employed, and stochastic gradient descent (SGD) optimization is used to
update the model parameters. Following model training, the script evaluates the trained
model on the validation subset. It computes metrics such as loss and accuracy to assess
the model's performance. Additionally, it calculates the F1 score for each emotion class
and the average F1 score, providing insights into the model's ability to correctly classify
different emotions.

The script concludes by visualizing the training and validation loss/accuracy curves over
epochs. These visualizations aid in understanding the model's training dynamics,
identifying overfitting or underfitting, and guiding further model refinement and optimization.

\begin{table}[h]
\centering
\begin{tabular}{|l|c|}
\hline
\text{Emotion} & \text{F1 Score} \\
\hline
\text{Neutral} & 0.4758 \\
\text{Anger} & 0.0963 \\
\text{Disgust} & 0.0099 \\
\text{Fear} & 0.0608 \\
\text{Happiness} & 0.3466 \\
\text{Sadness} & 0.1179 \\
\text{Surprise} & 0.1658 \\
\text{Other} & 0.3545 \\
\hline
\text{Average} & 0.2035 \\
\hline
\end{tabular}
\caption{F1 Scores for Different Emotions}
\label{tab:f1_scores}
\end{table}

In the training loss curve, a noticeable downward trend is observed, indicating a
reduction in the model's loss on the training dataset with each epoch. This suggests that
the model is learning to better fit the training data over time. Similarly, the training
accuracy curve exhibits an upward trend, demonstrating an improvement in the model's
ability to classify emotions correctly on the training data. Conversely, the validation loss
curve shows a fluctuating pattern, with occasional increases in loss across epochs. This
suggests that while the model may perform well on the training data, its performance on
unseen validation data fluctuates and does not consistently improve.The validation
accuracy curve shows a slight increase initially but plateaus, thereafter, indicating that
the model's performance on validation data remains relatively stable but does not
improve significantly with additional training epochs. Overall, these curves provide
insights into the training dynamics and generalization capabilities of the model.

\subsubsection{Action Unit Detection}
This problem focuses on detecting the activation of specific action units (AUs).
Annotation files provided in Train Set and Validation Set sub-folders contain commaseparated AU annotation values for each video frame, with AU activation represented by
1 and non-activation by 0. Frames with annotation values of -1 should be disregarded.
Action Unit Detection focuses on the detection of 12 specific Action Units (AUs) from
video data. The task comprises 547 videos, each annotated with 12 AUs, namely AU1,
AU2, AU4, AU6, AU7, AU10, AU12, AU15, AU23, AU24, AU25, and AU26. Notably,
seven of these videos feature two subjects, both of whom have been annotated. In total,
2,603,921 frames have been annotated in a semi-automatic procedure, involving a
combination of manual and automatic annotations. The annotation process occurs on a
frame-by-frame basis, ensuring comprehensive coverage of AUs across the dataset.

\begin{table}[h]
\centering
\begin{tabular}{|c|c|}
\hline
  Action Unit \# & Action   \\   \hline 
    \hline    
   AU 1 & inner brow raiser  \\   \hline 
   AU 2 & outer brow raiser  \\   \hline  
   AU 4 & brow lowerer  \\   \hline 
   AU 6 & cheek raiser  \\   \hline  
   AU 7 & lid tightener \\   \hline 
   AU 10 & upper lip raiser \\   \hline 
   AU 12 & lip corner puller \\   \hline 
   AU 15 & lip corner depressor \\   \hline 
   AU 23 & lip tightener \\   \hline 
   AU 24 & lip pressor  \\   \hline 
   AU 25 & lips part  \\   \hline 
   AU 26 & jaw drop  \\   \hline 
\end{tabular}
\end{table}

The performance of models in the Action Unit Detection task is evaluated based
on the average F1 Score across all 12 AUs. The F1 Score serves as the performance
measure, calculated as a weighted average of precision and recall. A high F1 Score,
ranging from 0 to 1, is desirable, indicating accurate detection performance.
The general formula being:

\begin{equation} \label{au}
\mathcal{P}_{AU} = \frac{\sum_{au} F_1^{au}}{12}
\end{equation}

\subsection{The Implementation}
This implementation presents a detailed overview of the Jupiter notebook for facial action
unit detection using deep learning techniques. It uses PyTorch, a popular deep learning
framework, along with other libraries for data manipulation and visualization. The script
begins by importing essential libraries required for various tasks throughout the pipeline.
These include libraries for file operations, mathematical computations, image
processing, deep learning model building (PyTorch), and data visualization. Paths to the
folders containing image data and annotation files are specified. Additionally, a
transformation pipeline is defined using PyTorch's transforms.Compose() module. This
pipeline includes operations such as resizing, random horizontal flipping, and random
rotation, aimed at augmenting the dataset for improved model generalization. A custom
dataset class (CustomDataset) is implemented to facilitate loading of image data and
corresponding annotations. This class parses annotation files to extract annotations for
each image, associating them with the respective image file paths. This ensures
seamless integration of image data and annotations for model training. Training and
validation datasets are instantiated using the custom dataset class. These datasets are
then converted into data loaders, which enable efficient batch-wise loading of data during
training and validation stages. Data loaders are crucial for optimizing model training
process and memory usage. Random samples from both training and validation datasets
are visualized to provide insights into the data distribution and quality. This step aids in
understanding the characteristics of the dataset and ensures proper data preprocessing.

The distribution of action units across the dataset is calculated and visualized using
Matplotlib. This analysis helps in understanding the prevalence of different action units
within the dataset, which is essential for model training and evaluation:
Total number of samples in the training dataset: 1343938
Total number of samples in the validation dataset: 439248
Sample Information:
Number of Action Units: 12

Subsets of the training and validation datasets are created for faster training and
experimentation purposes. By selecting a fraction of the original dataset, smaller subsets
are generated, reducing computational requirements while retaining representativeness.

A pre-trained ResNet18 model, is loaded using PyTorch's models module. The fully
connected layer of the model is modified to match the number of output features required
for the action unit detection task. The modified ResNet18 model is trained using a
weighted loss function and an Adam optimizer. A scheduler is employed to dynamically
adjust the learning rate during training based on the validation loss. The training process
is monitored using tqdm for progress visualization. The trained model is evaluated using
F1 score, a metric commonly used for binary classification tasks. F1 scores are
computed for individual action units as well as the average F1 score across all action
units. This evaluation provides insights into the model's performance and its ability to
detect various action units accurately

\begin{table}[h]
\centering
\begin{tabular}{|l|c|}
\hline
\text{Action Unit} & \text{F1 Score} \\
\hline
\text{Average} & 0.4003 \\
\text{AU1} & 0.4116 \\
\text{AU2} & 0.3454 \\
\text{AU4} & 0.3375 \\
\text{AU6} & 0.5097 \\
\text{AU7} & 0.6576 \\
\text{AU10} & 0.6522 \\
\text{AU12} & 0.6044 \\
\text{AU15} & 0.0279 \\
\text{AU23} & 0.0994 \\
\text{AU24} & 0.1222 \\
\text{AU25} & 0.7972 \\
\text{AU26} & 0.2379 \\
\hline
\end{tabular}
\caption{F1 Scores for Different Action Units}
\label{tab:au_f1_scores}
\end{table}

Throughout the training process, the model's performance is evaluated based on its
ability to minimize the loss function, which measures the disparity between predicted and
true values.
In this graph, we observe a gradual decrease in both training and validation losses
across the epochs, indicating that the model is learning and generalizing well to unseen
data. The training loss decreases from 0.2142 in the first epoch to 0.2046 in the final epoch, 
demonstrating consistent improvement in the model's ability to fit the training
data. Similarly, the validation loss decreases from 0.3907 to 0.3924 over the same
period, indicating that the model is not overfitting to the training data and is capable of
generalizing to unseen examples.

However, the validation loss slightly fluctuates throughout training, which could be
attributed to the inherent variance in the dataset or model sensitivity to specific data
samples. Despite these fluctuations, the overall trend of decreasing loss values suggests
that the model is learning effectively and making progress in optimizing its parameters
to better capture the underlying patterns in the data.

In this code, several strategies are employed to improve the accuracy of facial action unit
detection. Firstly, data augmentation techniques are utilized during training, including
random horizontal flipping and rotation of images. These transformations increase the
diversity of the training data, helping the model generalize better to unseen variations in
facial expressions. Additionally, a weighted loss function is implemented, assigning
higher weights to less frequent action units. This approach ensures that the model pays
more attention to rare action units during training, thereby improving its ability to detect
them accurately. Furthermore, a pre-trained ResNet18 model is leveraged as the
backbone architecture. Transfer learning from pre-trained models allows the model to
benefit from features learned on a large dataset, leading to improved performance even
with limited labelled data. The use of learning rate scheduling further enhances training
stability by dynamically adjusting the learning rate based on validation loss, preventing
the model from getting stuck. Lastly, the evaluation metric employed, F1 score, provides
a balanced measure of both precision and recall, offering a more accurate assessment
of the model's performance across all action units. Overall, through these strategies, the
code ensures comprehensive training, efficient learning, and robust evaluation,
ultimately leading to enhanced accuracy in facial action unit detection.

\subsubsection{Multi-Task-Learning implementation}
This is an Muti task approach to Facial expression recognition. This is a crucial task in
computer vision, with applications ranging from human-computer interaction to emotion
analysis. In this report, we present a multi-task learning approach to address three key
tasks in facial expression analysis: action unit detection, expression recognition, and
valence-arousal estimation. This model leverages a pre-trained ResNet-18 backbone to
extract features from facial images, followed by task-specific branches for each of the
three tasks.

The aim was to do all the unit tasks with one modified resent 18 model and compare
their performance.

The annotations are parsed from annotation files associated with each image.
Additionally, a transformation pipeline is defined for data augmentation during training,
including resizing, random horizontal flipping, and rotation.
A custom dataset class (CustomDataset) is created to organize the data and
annotations. It parses the annotation files and associates them with the corresponding
image paths. This larger custom dataset is made with the help of combining our previous
daalders into one.
Like the uni tasks, I have also created a random image generator, however this time is
gives us three different annotations for the same image. This way we know the data
loaders are working correctly.
A subset of the dataset is created for faster testing and debugging. The multi-task model
is defined, comprising a ResNet-18 backbone followed by task-specific fully connected
layers. The ResNet-18 backbone is loaded with pre-trained weights to extract features
from facial images. Task-specific branches are added for expression recognition, action
unit detection, and valence-arousal estimation. The training loop iterates over batches of
images and their associated annotations. For each batch, the model performs a forward
pass to compute the predictions for each task. Losses are computed for each task using
appropriate loss functions such as cross-entropy loss for expression recognition and
mean squared error loss for action unit detection and valence-arousal estimation. These
losses are backpropagated through the network, and model parameters are updated
using the Adam optimizer. Like the training loop, the validation loop evaluates the
model's performance on a separate validation dataset. It computes validation losses for
each task and updates the validation progress bar. The average validation loss for each
task is calculated at the end of each epoch

\begin{itemize}
    \item AEpoch 9/10, Train Expression Loss: 0.0001, Train AU Loss: 0.0000, Train VA Loss: 0.0000, Val Expression Loss: 0.1630, Val AU Loss: 0.0134, Val VA Loss: 0.0101
    \item .....
    \item Epoch 10/10, Train Expression Loss: 0.0000, Train AU Loss: 0.0000, Train VA Loss: 0.0000, Val Expression Loss: 0.0002, Val AU Loss: 0.0038, Val VA Loss: 0.0046
    \item Validation F1 Score for Expression Task: 0.2208
    \item Validation F1 Score for Action Unit Detection Task: 0.0865
    \item Average Validation CCC: 0.0274
\end{itemize}

In the training process, the model undergoes iterative improvements through ten epochs,
where it learns to predict expressions, action units, and valence-arousal estimates.
Despite initial fluctuations, the model converges, demonstrating improved performance
with minimized losses on both training and validation sets.

However, evaluation metrics, such as F1 score for expression recognition and
Concordance Correlation Coefficient for valence-arousal estimation, indicate some level
performance. 

After training, evaluation metrics are calculated to assess the model's performance. F1
score is computed for expression recognition and action unit detection tasks, while the
Concordance Correlation Coefficient (CCC) is calculated for valence-arousal estimation.
These metrics provide insights into the model's ability to generalize to unseen data and
capture correlations between predicted and ground truth values.

\subsubsection{Conclusion}
Throughout this project, we delved into three fundamental uni-task problems pivotal in
affect recognition: Valence-arousal estimation, Expression Recognition Classification,
and Action Unit Detection. These tasks are vital in understanding human emotions and
have wide-ranging applications across various domains, including human-computer
interaction, healthcare, and marketing. Valence-arousal estimation focuses on gauging
the emotional valence and arousal levels of individuals, providing insights into their
affective states. Expression Recognition Classification involves categorizing facial
expressions into predefined categories such as anger, happiness, and sadness,
enabling automated emotion analysis from facial images. Action Unit Detection, on the
other hand, aims to detect specific facial muscle movements, known as action units,
indicative of different emotions or affective states.
In addressing each task, we followed a systematic methodology encompassing data
preprocessing, model architecture design, training process, and evaluation metrics. Data
preprocessing involved meticulously preparing the datasets, handling annotations, and
augmenting the data for improved model generalization. We employed pre-trained
ResNet-18 models as the backbone architecture for feature extraction, fine-tuning them
for specific tasks. The training process involved iterative optimization using appropriate
loss functions, such as mean squared error for regression tasks and cross-entropy loss
for classification tasks. Evaluation metrics such as Concordance Correlation Coefficient
(CCC) for Valence-arousal estimation, F1 score for Expression Recognition
Classification, and F1 score for Action Unit Detection were utilized to assess model
performance.

In addressing each task, we followed a systematic methodology encompassing data
preprocessing, model architecture design, training process, and evaluation metrics. Data
preprocessing involved meticulously preparing the datasets, handling annotations, and
augmenting the data for improved model generalization. We employed pre-trained
ResNet-18 models as the backbone architecture for feature extraction, fine-tuning them
for specific tasks. The training process involved iterative optimization using appropriate
loss functions, such as mean squared error for regression tasks and cross-entropy loss
for classification tasks. Evaluation metrics such as Concordance Correlation Coefficient
(CCC) for Valence-arousal estimation, F1 score for Expression Recognition
Classification, and F1 score for Action Unit Detection were utilized to assess model
performance.

Across the three tasks, this project revealed several key findings and insights. We
observed that data augmentation techniques, such as random horizontal flipping and
rotation, significantly enhanced the model's ability to generalize to unseen data. The use
of weighted loss functions and class balancing techniques proved effective in addressing
class imbalances and improving model performance. Training dynamics, as evidenced
by loss and accuracy curves, showcased the models' ability to learn and adapt over
multiple epochs. Despite fluctuations in validation metrics, the overall trend indicated a
convergence towards improved model performance, highlighting the effectiveness of the
proposed methodologies.

In comparing the performance of the models across tasks, we found variations in model
proficiency and effectiveness. While some models exhibited higher accuracy and F1
scores, others faced challenges in accurately capturing certain affective nuances.
Performance comparisons with existing literature or benchmarks provided valuable
insights into the strengths and limitations of the proposed methodologies. Areas of
improvement, such as fine-tuning hyperparameters or exploring alternative model.

This project contributes to the ongoing discourse in affect recognition research by
proposing new methodologies and insights for addressing key uni-task problems. Our
findings advance our understanding of the intricacies involved in modeling human
emotions from facial expressions and pave the way for future research directions.
However, it's important to acknowledge the limitations of our project, including
constraints in dataset availability, computational resources, and model complexity.

In conclusion, this project points out the importance of affect recognition in understanding
human behavior and fostering advancements in technology and society. By leveraging
deep learning techniques and innovative methodologies, we've made progress in
addressing uni-task problems crucial in affect recognition.

Overall, I think we have completed the objectives and answered the research questions
in this project.

\textbf{Disclaimer:} Figures taken from:
\begin{itemize}
    \item \url{https://doi.org/10.48550/arXiv.2202.10659}
    \item \url{https://doi.org/10.48550/arXiv.1910.11090}
    \item \url{https://doi.org/10.48550/arXiv.2103.15792}
\end{itemize}

\textbf{Resources used:}
\begin{itemize}
    \item \url{https://doi.org/10.48550/arXiv.2105.03790}
    \item \url{https://doi.org/10.48550/arXiv.2303.00180}
    \item \url{https://doi.org/10.48550/arXiv.2105.0379}
    \item \url{https://pytorch.org/tutorials/beginner/data_loading_tutorial}
\end{itemize}

\bibliographystyle{ieee}
\bibliography{references}

\begin{thebibliography}{10}\itemsep=-1pt

\bibitem{arsenos2023data}
A.~Arsenos, A.~Davidhi, D.~Kollias, P.~Prassopoulos, and S.~Kollias.
\newblock Data-driven covid-19 detection through medical imaging.
\newblock In {\em 2023 IEEE International Conference on Acoustics, Speech, and Signal Processing Workshops (ICASSPW)}, pages 1--5. IEEE, 2023.

\bibitem{gerogiannis2024covid}
D.~Gerogiannis, A.~Arsenos, D.~Kollias, D.~Nikitopoulos, and S.~Kollias.
\newblock Covid-19 computer-aided diagnosis through ai-assisted ct imaging analysis: Deploying a medical ai system.
\newblock {\em arXiv preprint arXiv:2403.06242}, 2024.

\bibitem{hu2024bridging}
G.~Hu, E.~Papadopoulou, D.~Kollias, P.~Tzouveli, J.~Wei, and X.~Yang.
\newblock Bridging the gap: Protocol towards fair and consistent affect analysis.
\newblock {\em arXiv preprint arXiv:2405.06841}, 2024.

\bibitem{kollias2022abaw}
D.~Kollias.
\newblock Abaw: Learning from synthetic data \& multi-task learning challenges.
\newblock In {\em European Conference on Computer Vision}, pages 157--172. Springer, 2022.

\bibitem{kollias2023multi}
D.~Kollias.
\newblock Multi-label compound expression recognition: C-expr database \& network.
\newblock In {\em Proceedings of the IEEE/CVF Conference on Computer Vision and Pattern Recognition}, pages 5589--5598, 2023.

\bibitem{kollias2023ai}
D.~Kollias, A.~Arsenos, and S.~Kollias.
\newblock Ai-enabled analysis of 3-d ct scans for diagnosis of covid-19 \& its severity.
\newblock In {\em 2023 IEEE International Conference on Acoustics, Speech, and Signal Processing Workshops (ICASSPW)}, pages 1--5. IEEE, 2023.

\bibitem{kollias2023deep2}
D.~Kollias, A.~Arsenos, and S.~Kollias.
\newblock A deep neural architecture for harmonizing 3-d input data analysis and decision making in medical imaging.
\newblock {\em Neurocomputing}, 542:126244, 2023.

\bibitem{kollias2024domain}
D.~Kollias, A.~Arsenos, and S.~Kollias.
\newblock Domain adaptation, explainability \& fairness in ai for medical image analysis: Diagnosis of covid-19 based on 3-d chest ct-scans.
\newblock {\em arXiv preprint arXiv:2403.02192}, 2024.

\bibitem{kolliasijcv}
D.~Kollias, S.~Cheng, E.~Ververas, I.~Kotsia, and S.~Zafeiriou.
\newblock Deep neural network augmentation: Generating faces for affect analysis.
\newblock {\em International Journal of Computer Vision}, pages 1--30, 2020.

\bibitem{kollias2023facernet}
D.~Kollias, A.~Psaroudakis, A.~Arsenos, and P.~Theofilou.
\newblock Facernet: a facial expression intensity estimation network.
\newblock {\em arXiv preprint arXiv:2303.00180}, 2023.

\bibitem{kollias2020analysing}
D.~Kollias, A.~Schulc, E.~Hajiyev, and S.~Zafeiriou.
\newblock Analysing affective behavior in the first abaw 2020 competition.
\newblock In {\em 2020 15th IEEE International Conference on Automatic Face and Gesture Recognition (FG 2020)(FG)}, pages 794--800.

\bibitem{kollias2019face}
D.~Kollias, V.~Sharmanska, and S.~Zafeiriou.
\newblock Face behavior a la carte: Expressions, affect and action units in a single network.
\newblock {\em arXiv preprint arXiv:1910.11111}, 2019.

\bibitem{kollias2021distribution}
D.~Kollias, V.~Sharmanska, and S.~Zafeiriou.
\newblock Distribution matching for heterogeneous multi-task learning: a large-scale face study.
\newblock {\em arXiv preprint arXiv:2105.03790}, 2021.

\bibitem{kollias2024distribution}
D.~Kollias, V.~Sharmanska, and S.~Zafeiriou.
\newblock Distribution matching for multi-task learning of classification tasks: a large-scale study on faces \& beyond.
\newblock {\em arXiv preprint arXiv:2401.01219}, 2024.

\bibitem{kollias2023abaww}
D.~Kollias, P.~Tzirakis, A.~Baird, A.~Cowen, and S.~Zafeiriou.
\newblock Abaw: Valence-arousal estimation, expression recognition, action unit detection \& emotional reaction intensity estimation challenges.
\newblock In {\em Proceedings of the IEEE/CVF Conference on Computer Vision and Pattern Recognition}, pages 5888--5897, 2023.

\bibitem{kollias20246th}
D.~Kollias, P.~Tzirakis, A.~Cowen, S.~Zafeiriou, C.~Shao, and G.~Hu.
\newblock The 6th affective behavior analysis in-the-wild (abaw) competition.
\newblock {\em arXiv preprint arXiv:2402.19344}, 2024.

\bibitem{kollias2019deep}
D.~Kollias, P.~Tzirakis, M.~A. Nicolaou, A.~Papaioannou, G.~Zhao, B.~Schuller, I.~Kotsia, and S.~Zafeiriou.
\newblock Deep affect prediction in-the-wild: Aff-wild database and challenge, deep architectures, and beyond.
\newblock {\em International Journal of Computer Vision}, 127(6):907--929, 2019.

\bibitem{kollias2023btdnet}
D.~Kollias, K.~Vendal, P.~Gadhavi, and S.~Russom.
\newblock Btdnet: A multi-modal approach for brain tumor radiogenomic classification.
\newblock {\em Applied Sciences}, 13(21):11984, 2023.

\bibitem{kollias2019expression}
D.~Kollias and S.~Zafeiriou.
\newblock Expression, affect, action unit recognition: Aff-wild2, multi-task learning and arcface.
\newblock {\em arXiv preprint arXiv:1910.04855}, 2019.

\bibitem{kollias2021affect}
D.~Kollias and S.~Zafeiriou.
\newblock Affect analysis in-the-wild: Valence-arousal, expressions, action units and a unified framework.
\newblock {\em arXiv preprint arXiv:2103.15792}, 2021.

\bibitem{kollias2021analysing}
D.~Kollias and S.~Zafeiriou.
\newblock Analysing affective behavior in the second abaw2 competition.
\newblock In {\em Proceedings of the IEEE/CVF International Conference on Computer Vision}, pages 3652--3660, 2021.

\bibitem{psaroudakis2022mixaugment}
A.~Psaroudakis and D.~Kollias.
\newblock Mixaugment \& mixup: Augmentation methods for facial expression recognition.
\newblock In {\em Proceedings of the IEEE/CVF Conference on Computer Vision and Pattern Recognition}, pages 2367--2375, 2022.

\bibitem{zafeiriou2017aff}
S.~Zafeiriou, D.~Kollias, M.~A. Nicolaou, A.~Papaioannou, G.~Zhao, and I.~Kotsia.
\newblock Aff-wild: valence and arousal'in-the-wild'challenge.
\newblock In {\em Proceedings of the IEEE conference on computer vision and pattern recognition workshops}, pages 34--41, 2017.

\end{thebibliography}

\end{document}